\documentclass[11pt]{article}
\pdfoutput=1
\usepackage{verbatim}
\usepackage{subfigure}
\usepackage{natbib}
\usepackage{amsfonts}
\usepackage{amsmath}
\usepackage[mathscr]{eucal}
\usepackage{graphicx}
\usepackage{setspace}
\bibpunct{(}{)}{;}{a}{,}{,}
\renewcommand{\c}[1]{\noindent \emph{***#1}}
                                                                                                      \renewcommand{\c}[1]{\noindent \emph{***#1}}

                                                        \bibpunct{(}{)}{;}{a}{,}{,}
                                                        \renewcommand{\c}[1]{\noindent \emph{***#1}}

\title{The Fundamental Problem with the \\ Building Block Hypothesis}
\author{Keki Burjorjee\\Computer Science Department,\\Brandeis University, Waltham 02454.\\kekib@cs.brandeis.edu}
\date{}
\begin{document}

\maketitle


                                                                                                                    \begin{abstract}
                                                                                                                    Skepticism of the building block hypothesis (BBH) has previously been expressed on account of the weak theoretical foundations of this hypothesis and the anomalies in the empirical record of the simple genetic algorithm. In this paper we hone in on a more fundamental cause for skepticism---the extraordinary strength of some of the assumptions that undergird the BBH. Specifically, we focus on assumptions made about the distribution of fitness over the  genome set, and argue that these assumptions are unacceptably strong. As most  of these assumptions have been embraced by the designers of so-called ``competent" genetic algorithms, our critique is relevant to an appraisal of such algorithms as well.\\[.3cm]
                                                                                                                    \end{abstract}

                                                                                                                    \section{Introduction}
                                                                                                                    In constructing a representation (a genome-to-phenotype map and a fitness function) a GA practitioner implicitly determines how fitness gets distributed over a genome set. If a GA with this representation is adaptive then, with overwhelmingly high probability, the induced fitness distribution has some type of ``structure" that the GA is exploiting. There can be no other reason for the GA's performance. GAs are frequently adaptive in practice. This entails that GA practitioners often construct representations that induce fitness distributions with GA-exploitable structure.

                                                                                                                    Before proceeding further, let us clarify what me mean by the word ``adaptive": given a search problem, we say that some population based search algorithm is \emph{adaptive} if,  across several runs with different random seeds, the average fitness of the population consistently trends upwards. By this token GAs are often adaptive in practice, whereas population based random search is not. We refer to this feature of genetic algorithms as their \emph{adaptive capacity}.

                                                                                                                    We posit that \emph{any}coherent theory about the adaptive capacity of genetic algorithms must consist of the following: Firstly a set of assumptions about the way fitness commonly gets distributed via the representational choices of GA practitioners; we call these the \emph{commonplace fitness structure assumptions} (CFSAs). And secondly, a hypothesis about how this fitness structure gets exploited by a GA during adaptation; we call this the \emph{exploitation hypothesis} (EH).

                                                                                                                   The exploitation hypothesis depends critically on the commonplace fitness structure assumption, but not vice-versa. The CFSAs, in other words, are foundational.  Therefore, when developing an explanation for the adaptive capacity of genetic algorithms, getting the CFSAs right is extremely important. Fundamentally flawed CFSAs thwart the entire enterprise no matter how much effort is lavished upon the development and justification of the EH.

                                                                                                                   Practically any EH can be justified if one starts with sufficiently strong assumptions. To be viable an EH must be based upon CFSAs that are \emph{weak}. This is nothing but an application of the principle of Occam's razor, which holds that the weaker the assumptions that undergird a theory, the more viable the theory. This principle clearly makes sense in the current context. GA representations are completely ad-hoc. Therefore the weaker our assumptions about the nature of the induced fitness structure, the more likely it is that these assumptions hold true\footnote{Adherents of the building block hypothesis may disagree with our assessment that the representations they construct are ``completely ad-hoc". After all, much advice for constructing representations has been dispensed (e.g. ``ensure a large supply of  building blocks"); those who have made an effort to follow this advice may claim to have a basis for making strong assumptions about the structure of the fitness distributions they induce. Unfortunately claims of this nature are unjustified. While there is plenty of advice on how representations should be constructed, there is, as far as we can tell, no principled way for determining how this advice should be put into practice in specific instances (except perhaps when the problems are contrived). In this respect the dispensed advice is much like the famous investing maxim ``buy low, sell high"---easy to state, hard to implement.}.

                                                                                                                   The building block hypothesis (BBH) is currently the dominant explanation for the adaptive capacity of simple genetic algorithms. Though this hypothesis has come in for some (at times sharp) criticism in recent times, it remains the compass by which the vast majority of GA practitioners---past and present---have allowed themselves to be guided. It is also the first explanation for the adaptive capacity of GAs that most students receive, and in this capacity surely exerts an anchoring effect \citep{tversky1974juu} on their reasoning.

                                                                                                                   Previously expressed skepticism of the building block hypothesis can be divided into two categories. The first consists of criticism of the weak theoretical foundations of this hypothesis  (for a survey see \citealt[section 3.3]{reeves2003gap}). Proponents of the building block hypothesis have, for the most, part brushed aside criticism of this sort. Goldberg, for example, calls such critiques  ``[a] favorite parlor game in genetic and evolutionary computation circles"\citep[p7]{desInnov}, and, by way of analogy, characterizes such concerns as absurd. ``[T]he very idea that an airplane is ineffective or unsafe because a formal mathematical proof of flight does not exist is itself an absurdity.", he writes.  ``Yet, if this is so---and no proof does exist of airplane flight---and if I \ldots transform the aircraft into a \emph{genetic algorithm}\ldots, why is it that [this] patently absurd alarm seems so real in the context of GAs and their design and use?"\citep[p19]{desInnov}.

                                                                                                                   The second category is comprised of skepticism arising from the anomalous performance of the simple genetic algorithm on some basic empirical tests (\citealt{forrest93relative}; \citealt[Section 6.2]{watsonBook}). In response to these results proponents of the building block hypothesis have downplayed the importance of the simple genetic algorithm \citep{journals/ec/Holland00, desInnov}, and have advocated the use other sorts of  genetic algorithms (e.g. cohort genetic algorithms, ``competent" genetic algorithms), that are more complicated than the simple genetic algorithm, and typically contain mechanisms that are not biologically plausible. Strictly speaking the building block hypothesis applies only to the simple genetic algorithm. Therefore by downgrading the importance of the simple genetic algorithm proponents of the building block hypothesis can claim to have rendered skepticism about the veracity of this hypothesis irrelevant.

                                                                                                                    The skepticism expressed in this paper does not belong to either of the two categories described above. It is, in a sense, more \emph{fundamental}, stemming from a critical appraisal of the strength of the CFSAs that undergird the building block hypothesis. We examine various influences---historical, social, and metaphysical---that have shaped these assumptions, and argue that the resulting CFSAs are unacceptably strong. As these CFSAs have largely been embraced by the designers of the new types of genetic algorithms mentioned in the previous paragraph, our criticism is also relevant to an evaluation of those algorithms.

                                                                                                                    The rest of this paper is organized as follows. In section \ref{history} we briefly recount the history of the building block hypothesis---its origin, ascent, and recent troubles. In section \ref{CFSAs} we describe the CFSAs undergirding this hypothesis, and explain why we find them to be unacceptably strong. In sections \ref{appeal1} and \ref{appeal2} we critically examine the ways in which proponents of the building block hypothesis have sought to justify this hypothesis, and by extension, the CFSAs that undergird it. In section \ref{looselinkage} we explain the import of our criticism of these CFSAs for the current direction of the field of genetic algorithmics. And finally as part of our conclusion, we draw parallels between the CFSAs of the building block hypothesis, and the now defunct concept of luminiferous aether that was popular in nineteenth century theories about the propagation of light.

                                                                                                                    \section{A Brief History of the BBH}\label{history}
                                                                                                                    Scientific theories are typically presented without reference to the context within which they were developed \citep[p79]{okasha2002psv}. At a certain level this of course makes sense; surely what a scientist has for breakfast is immaterial to one's evaluation of her theories. At the same time, it has been observed that to genuinely understand the state of a science, an acquaintance with its history is essential. Every scientific theory is undergirded by what the historian of science, Thomas Kuhn, calls \emph{received beliefs} \citep[p4]{kuhn:ssr}---assumptions transmitted from one generation to the next within a scientific community. An acquaintance with the history of a science can help one identify the origins of such assumptions and the circumstances under which they have have been perpetuated.

                                                                                                                    \subsection{Origin of the Building Block Hypothesis}

                                                                                                                    In the 1960s and early  '70s  Holland developed an abstract mathematical model of adaptive processes, what he called an \emph{adaptive plan}, which was inspired by natural evolution but was of greater generality. Holland sought to use this model to unify, under one theoretical framework, adaptation  in such diverse fields as neuroscience, economics, control, game theory, artificial intelligence, and genetics.

                                                                                                                    For any adaptive process that generates only valid structures of some type, one can define a set $\mathscr A$ of all possible valid structures that may be generated during the adaptive process. For example, in the domain of economic planning an element of $\mathscr A$ may be a mix of goods; in game-theory $\mathscr A$ may be the set of all strategies with respect to some game \citep[p4]{holland75:_adapt_natur_artif_system}. Holland conceptualized adaptation as a process that generates samples from $\mathscr A$, concentrating these samples over time in subsets of $\mathscr A$ of increasing average fitness.

                                                                                                                    The central conceptual objects in Holland's framework are subsets of $\mathscr A$ called schemata (singular schema). Holland noted that each point in $\mathscr A$ may belong simultaneously to several schemata. Therefore an evaluation of the fitness of that point is, in effect, a fitness evaluation of a sample from each schema that the point belongs to. If the point is `fit' then this reflects well on all of those schemata, and vice-versa if the point is `unfit'. This observation suggested to Holland the possibility of the existence of algorithms which, by testing small numbers of points, implicitly test vast numbers of schemata, and then implicitly use this information to concentrate trials in schemata of increasing average fitness. Holland named this phenomenon \emph{intrinsic parallelism} \citep[p74]{holland75:_adapt_natur_artif_system}, a name he later revised to \emph{implicit parallelism} \citep{Holland92}. Holland also clearly seems to have been impressed by the utility of hierarchical assembly (\citealp[Chapter 4]{Simon69}; \citealp[p168]{holland75:_adapt_natur_artif_system}).

                                                                                                                    In an argument that freely mixed speculation and deduction---the line between the two was typically left blurry---Holland concluded that a specific adaptive plan that models natural evolution---what he called a \emph{genetic plan}---can generate high-fitness solutions to difficult adaptation problems, and that this adaptive plan will do so using implicit parallelism and hierarchical assembly.

                                                                                                                    Starting in 1970, Holland's students began applying implementations of genetic plans to adaptation problems (e.g. \citealp{cavicchio, Hollstien}). They found that these algorithms typically outperformed random search.  In a landmark dissertation De Jong \citeyearpar{DeJongDissertation} described experiments in which a stripped down version of the genetic plan---what is now called the simple genetic algorithm---was applied to a carefully contrived set of fitness functions with well-understood and diverse characteristics. De Jong \citeyearpar{DeJongDissertation} reports that ``Out of these early studies [his and those of his colleagues] emerged a picture of a GA as a robust adaptive search procedure, which was surprisingly effective as a global search heuristic".

                                                                                                                    Holland's theoretical work on genetic plans was seen as the obvious place to look for an explanation for the adaptive capacity of genetic algorithms. Holland and his students simplified this work and settled upon the well known explanation that goes by the name of the building  block hypothesis \citep{Goldberg:1989:GAS,Holland92,Mitchell:1996:IGA}.

                                                                                                                    \subsection{Initial Espousal and Recent Skepticism }
                                                                                                                    Until the late 1980s the building block hypothesis seems to have gone relatively unquestioned. However, with the explosive increase in the popularity of the genetic algorithm came increasing scrutiny of its theoretical foundations. Starting in the early 1990s several researchers began to publish independent ground-up theoretical analyses of the dynamics of genetic algorithms \citep{VoseLiepins91,nix1992mga,Vose1993,prugelbennett1994aga,RattrayThesis,shapiro:2001:smtga}. What prompted these entirely new lines of theoretical analysis? It is hard to say for certain, but we believe that at least part of the cause was frustration with the unclear demarcation between conjectures and mathematically provable facts that is characteristic of the argument of the argument for the building block hypothesis. In the preface to his book on the simple genetic algorithm Vose \citeyearpar{vose:1999:sgaft} writes ``My central purpose in writing this book is to provide an introduction to what is known about the theory of the Simple Genetic Algorithm. The rigor of mathematics is employed so as not to inadvertently repeat myths or recount folklore".  He adds that the absence of core elements of ``standard GA theory" in his book is due to the unintelligibility, the irrelevance, or the mathematical unjustifyability of these elements. In a later work Wright et. al. \citeyearpar{conf/gecco/WrightVR03} remark, ``The various claims about GAs that are traditionally made under the name of the \emph{building block hypothesis} have, to date, no basis in theory".

                                                                                                                   Statements like these have served a vital purpose. Despite its name, the building block hypothesis had come to be treated as much more than a hypothesis.  It had become the de-facto explanation for the success of genetic algorithms, thoroughly shaping the paradigm within which most GA research was conducted. For example, this hypothesis determined what constituted a valid question, a valid explanation, a valid prediction, and a valid enhancement of some genetic algorithm. Given the assertive tone in which the building block used to be presented (see \citealp{Goldberg:1989:GAS,Holland92}), and the blurry line between deductive reasoning and conjecture in the argument for this hypothesis, students and non-theoreticians who were eager for an explanation cannot not be blamed for surmising that the building block hypothesis is based largely on deductive reasoning. Statements such as the ones reproduced above now serve to caution them against this notion. These statements have also served as a call to apologists to clearly describe their premises and modes of reasoning. The responses elicited (e.g. \citealp{journals/ec/Holland00,desInnov}), provide us with the clearest picture yet of the presumptions undergirding the building block hypothesis.

                                                                                                                   \section{The CFSAs of the BBH\label{CFSAs}}

                                                                                                                    Let us quickly recount some basic elements of schema theory. In the case of genetic algorithms,  $\mathscr A$ is the set of all strings of some predetermined length drawn from some alphabet (in what follows we assume that this alphabet is \{0,1\}). Let us call the elements of this set \emph{genomes}. Schemata are represented by so called `similarity templates'. Suppose $\mathscr A$ is a set of strings of length 6,  then the schema 1*0**0 is the subset of  strings in $\mathscr A$ with 1 in the first position, zero in the third and sixth positions, and either 1 or 0 in the second, fourth, and fifth positions; the *, called a `wildcard', stands for `don't care'. For the sake of brevity, a schema template is often just called a schema. It is important though to keep in mind the distinction between the two. Given some population, the \emph{frequency} of a schema is the number of genomes in the population that belong to that schema. A \emph{defining position} of a schema is a position that is not a wildcard. The \emph{defining length} of a schema is the difference between the indices of the last and first defining positions. Finally, the \emph{order} of a schema is the number of defining positions. Thus, the defining length and order of the schema in the example above are five and three respectively. A schema with low defining-length (and therefore low order) is said to be `short'.

                                                                                                                   Let $S_1$ and $S_2$ be two subsets of  $\mathcal A$. For a population of size $N$, let us say that the sampling fitness of $S_1$ is likely to be greater than (or less than) the sampling fitness of $S_2$ if there is a high probability  that the average fitness of $\frac{|S_1|}{|\mathcal A|}N$ samples drawn uniformly from  $S_1$ will be greater than (or, respectively, less than) the average fitness of $\frac{|S_2|}{|\mathcal A|}N$ samples drawn uniformly from $S_2$.

\begin{figure}\begin{center}
                                                                                                                    \subfigure[A consonant intersection between two subsets]{
                                                                                                                    \includegraphics[height=4.8cm]{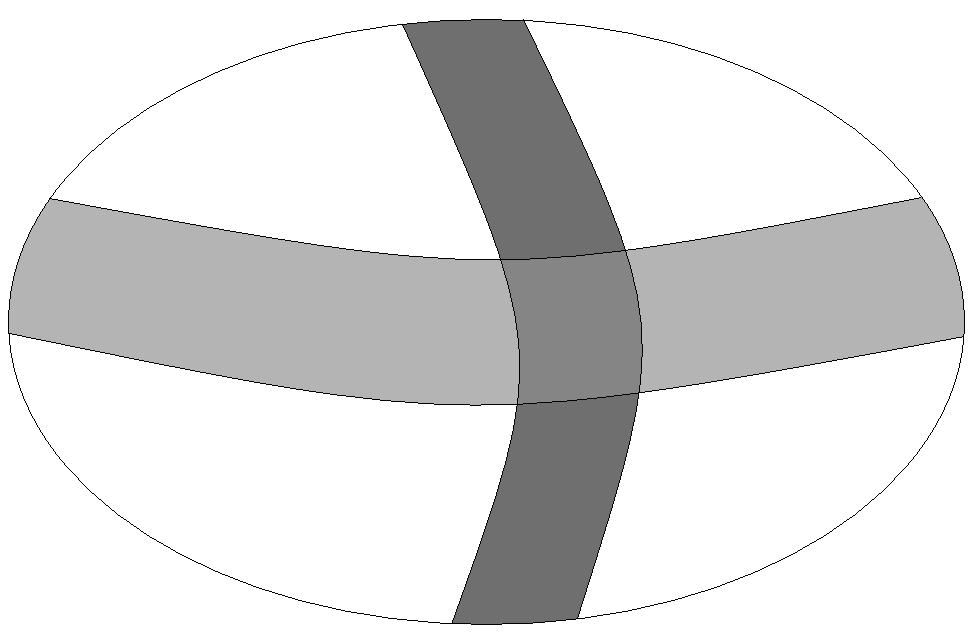}\quad\quad}\\
                                                                                                                    \subfigure[An antagonistic intersection between two subsets]{
                                                                                                                    \includegraphics[height=4.8cm]{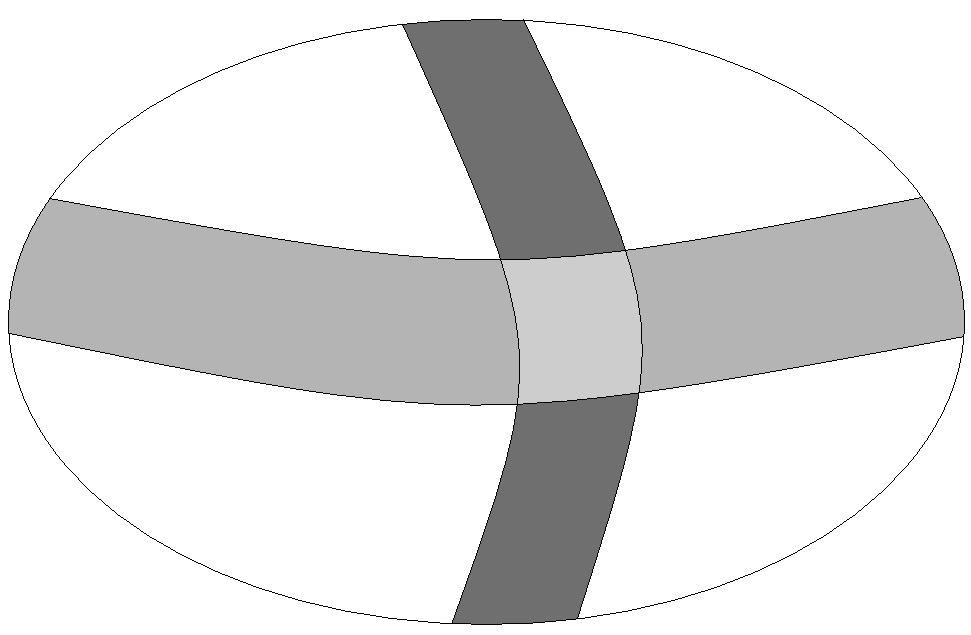}\quad\quad}\\
                                                                                                                    \subfigure[A synergistic intersection between two subsets]{
                                                                                                                    \includegraphics[height=4.8cm]{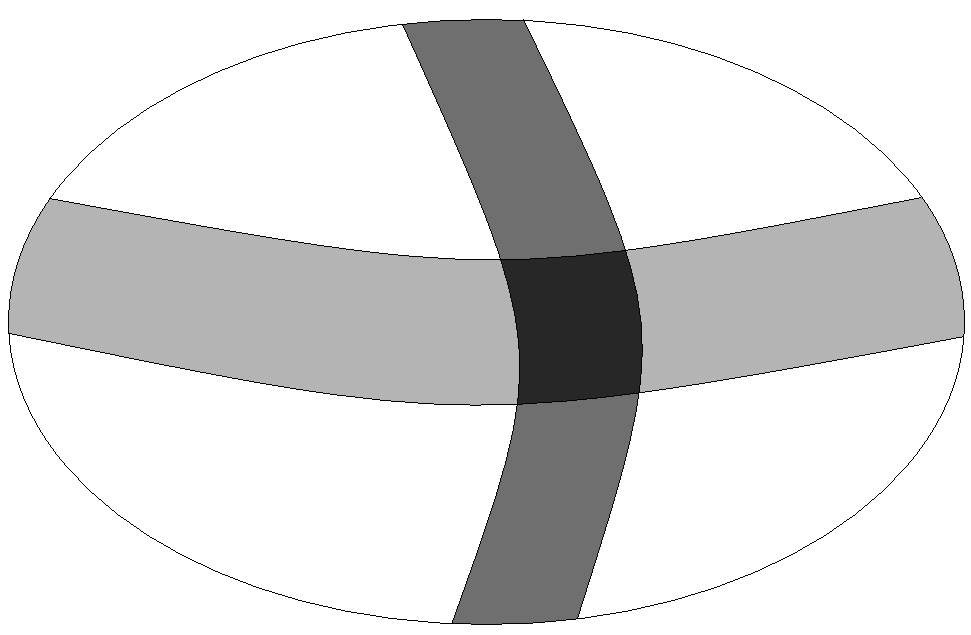}\quad\quad}
                                                                                                                    \end{center}
                                                                                                                    \caption{\label{conantsyn}A consonant, antagonistic, and synergistic intersection between two subsets is depicted. Darker shades signify greater sampling fitness}
                                                                                                                    \end{figure}

                                                                                                                   Given some collection of subsets of $\mathcal A$ with a non-empty intersection, we say that the intersection is \emph{consonant} (Figure \ref{conantsyn}a) if the sampling fitness of the intersection is not likely to be greater than or less than the sampling fitness of any of the intersecting subsets. We say that the intersection is \emph{antagonistic} (Figure \ref{conantsyn}b) if the sampling fitness of the intersection is likely to be less than the sampling fitness values of all the intersecting subsets.  And we say that the intersection is \emph{synergistic} (Figure \ref{conantsyn}c) if the sampling fitness of the intersection is likely to be greater than the sampling fitness values of all the intersecting subsets.

                                                                                                                   We define a \emph{basic building block} to be a short schema with sampling fitness that is likely to be greater than the uniform sampling fitness of $\mathcal A$.  A synergistic intersection between a small collection of basic building blocks is called a 2nd level building block, a synergistic intersection between a small collection of  2nd level building blocks is called a 3rd level building block, and so on. The building block hypothesis rests on (at the very least) the following two CFSAs:
                                                                                                                   \begin{description}
                                                                                                                   \item[Abundant Basic Building Blocks:]  A large number of basic building blocks exist.
                                                                                                                   \item [Heirarchical Synergism:] Antagonistic intersections between the building blocks of any level are rare, whereas synergistic intersections  between small collections of lower level  building blocks are common.
                                                                                                                    \end{description}

                                                                                                                   It seems to us that the number of ways in which these assumptions can be satisfied is vastly outnumbered by the number of ways in which they will not be satisfied. We trust that the reader, upon seeing these assumptions explicitly laid out, will agree that they are unacceptably strong. Remember that we are talking about the structure of fitness functions induced by the ad-hoc decisions of GA practitioners engaged in solving poorly understood, or NP-hard problems.

                                                                                                                   In the following sections we examine two ways in which proponents of the building block hypothesis have attempted to justify their belief in these assumptions---an appeal to certain metaphysical positions, and an appeal to authority.

                                                                                                                   \section{Appeal to Metaphysical Positions\label{appeal1}}

                                                                                                                   In support of the building block hypothesis Holland has asserted the \emph{building block thesis}.  He describes this thesis as follows: `The ``building block thesis"  holds that most of what we know about the world pivots on descriptions and mechanisms constructed from elementary building blocks'.  He characterizes building blocks as ``parts" such that ``(i) they must be easy to identify (once they've been discovered or picked out), and (ii) they must be readily recombined to form a wide variety of structures (much as can be done with children's building blocks)".

                                                                                                                    The way this Building block thesis is phrased it is a statement about what is already known about the world rather than a universal law, which is how Holland goes on to use it.  Our most sympathetic rephrasing of the building block thesis in light of the way Holland uses it to support the building block hypothesisis is as follows: Building blocks play a key role in the structure and function of most objects and processes. Building blocks are  (i) parts of wholes,  (ii) easily identifiable, and  (iii) recombinable with other building blocks to form a wide variety of forms. It is necessary to quote Holland at length so as not to leave out any part of his argument for this thesis. Holland writes
                                                                                                                   \begin{quote}
                                                                                                                    `The successive levels of building blocks used in physics are familiar to anyone interested in science---nucleons constructed from quarks, nuclei constructed from nucleons, atoms constructed from nuclei, molecules constructed from atoms, and so on \ldots Nowadays a similar succession presents itself in daily newspaper articles discussing progress in biology: chromosomal DNA constructed from 4 nucleotide building blocks; the basic structural components of enzymes: alpha helices, beta sheets, and the like, constructed from 20 amino acids; standard ``signalling" proteins for turning genes ``on" and ``off," ad ``autocatalytic bio-circuits," such as the citric acid cycle, that perform similar functions over extraordinarily wide ranges of species organelles constructed from situated bio-circuits, and so on \ldots And, of course, there are the long-standing taxonomic categories: species, genus, family, etc., specified in terms of morphological and chromosomal building blocks held in common. However the pervasiveness of building blocks only becomes apparent when we start looking at other areas of human endeavor. In some cases we take building blocks so much for granted that we're not even aware of them. Human perception is a case in point. The objects we recognize in the world are always defined in terms of elementary, reusable building blocks, be they trees (leaves, branches, trunks, \ldots), horses (legs, body, neck, head, blunt teeth, \ldots), speech(a limited set of basic sounds called phonemes), or written language (the 26 letters of English, for example).

                                                                                                                    `In other cases we just don't make the building blocks explicit. Consider two major inventions of the 20th century, the internal combustion engine and the electronic computer. The building blocks of the internal combustion engine: gears, Venturi's aspirator, Galvani's sparking device, and so on, were well-known prior to the invention. The invention consisted in combining them in a new way. Similarly, the components of early electronic programmable computers: wires, Geiger's counting device, cathode ray tubes, and the like, were well-known. Even earlier, Babbage had spelled out an overall architecture using long-standard, mechanical building blocks, (gears, ratchets, levers, etc.). The latter invention consisted in combining the electronic building blocks in a way that implemented Babbage's mechanical layout. And, of course, building blocks underpin the critical step for universal computation: arbitrary algorithms are constructed by combining copies of a small set of basic instructions. For both the internal combustion engine and the programmable computer, the building blocks were a necessary precursor, but the innovation required a new combination of the blocks' \citep{journals/ec/Holland00}.
                                                                                                                    \end{quote}

                                                                                                                    \citet{journals/ec/Holland00} regards the schema theorem as the Rosetta stone that shows how the building block thesis applies to genetic algorithms.  This theorem shows that if a short schema with frequency $x$ has above average fitness in some generation $t$, then the expected frequency of that schema in generation $t+1$ is greater than $x$. From this result proponents of the building block hypothesis conclude that short schemata with above average fitness are the basic building blocks that genetic algorithms implicitly use. That such building blocks must ``therefore" be abundant and hierarchically synergistic presumably  ``follows" from the building block thesis.

                                                                                                                    We trust that the reader will agree that the building block thesis is too vague to be \emph{falsifiable}. Firstly, it uses highly subjective language. The stipulations that building blocks be ``easy" to identify and ``recombinable"  beg the question ``according to whom?''. Secondly, the use of the word ``most" makes this thesis impossible to falsify unless one conducts an inventory of all entities in the universe.

                                                                                                                    The falsifiability criterion was formulated by the philosopher of science  Karl Popper \citeyearpar{logicofscientificdisc,conjrefut} as a way to distinguish between \emph{scientific} theories, such as Einstein's theory of gravitation, and \emph{pseudo-scientific} theories, such as astrology, or Freud's theory of psycho-analysis. Pseudo-scientific theories, noted Popper, \begin{quote}``appeared to be able to explain practically everything that happened within the fields to which they referred. The study of any of them seemed to have the effect of an intellectual conversion or revelation, opening your eyes to a new truth hidden from those not yet initiated. Once your eyes were thus opened you saw confirming instances everywhere: the world was full of \emph{verifications} of the theory. Whatever happened always confirmed it. Thus its truth appeared manifest and unbelievers were clearly people who did not want to see the manifest truth"\citep[p45]{conjrefut}.\end{quote}Scientific theories, in contrast, are theories that take risks---by predicting unexpected phenomena (e.g. gravitational lensing) they leave themselves open to refutation.  ``A theory which is not refutable by any conceivable event is non-scientific", wrote Popper. ``Irrefutability is not a virtue (as people often think) but a vice"\citep[p46]{conjrefut}.

                                                                                                                    Consider for example the difference between the Church-Turing thesis \citep{Copeland}, a refutable and therefore scientific thesis, and the building block ``thesis", which is neither. The most generous way to regard the building block ``thesis" is as a metaphysical theory of pan-modularity and pan-hierarchism. Should this new way of regarding Holland's  ``thesis" allay our concerns about the building block hypothesis? It should not. Indeed \emph{any} hypothesis can be justified by first asserting a generalization of the hypothesis as a new metaphysical position, and then using the metaphysical position to argue in favor of the specific hypothesis.

                                                                                                                    \section{Appeal to Authority\label{appeal2}}

                                                                                                                     It is clear from his writings that Holland views the building block hypothesis as a straightforward generalization of Fisher's theory of adaption (\citeyear[p89]{holland75:_adapt_natur_artif_system}; \citeyear{journals/ec/Holland00})---a generalization indeed that, to Holland's mind,  rests on much \emph{weaker} assumptions than Fisher's. Fisher's theory currently reigns as the orthodox view in Population Genetics, so it is understandable that Holland finds it exasperating that the building block hypothesis should meet with the kind of criticism it has received from certain quarters within the genetic algorithmics community (see \citealp{journals/ec/Holland00}).

                                                                                                                   In this section we describe some of Fisher's assumptions, highlight their extraordinary strength, and  examine the circumstances under which these assumptions became part of the orthodoxy of Population Genetics. It is important to stress that the adoption of Fisher's assumptions by population geneticists, though pervasive, is by no means unanimous. Accordingly, we will review some of the criticism that has been leveled at these assumptions. Finally we compare Holland's assumptions with Fisher's. We argue that the extent by which the former are weaker than the latter has been exaggerated, and explain how Holland's assumptions are in fact stronger than Fisher's in an important respect.

                                                                                                                   \subsection{The Fisherian Pardigm and its Discontents}

                                                                                                                    For almost two decades after the discovery, in 1900, of his paper on inheritance in pea plants, Mendel's theory of particulate inheritance was thought to be at odds with the theory of adaptation by natural selection proposed by Darwin and Wallace. The \emph{mendelians} argued that new species arise not by gradual changes, but by large jumps---called saltations---caused by macromutations. Such mutations (which were assumed to occur infrequently) were thought to be the main drivers of adaptation. Natural selection on the other hand was thought to play at best a minor part---that of mopping up deleterious macromutations. Though this point of view was never articulated by Mendel himself, his name became associated with researchers such as DeVries, Bateson and Johansen who used the results of Mendel's paper to downplay the effects of natural selection (p 777 Mayr).

                                                                                                                    Opposing them, the \emph{biometricians} (e.g. Pearson, and Weldon) noted that gradualness abounds in nature, and argued that evolution consists of a gradual shift of an entire population rather than the creation of new types by macromutation. The biometricians made extensive use of statistics to study the effect of natural selection on phenotypic distributions and claimed superiority over the mendelians on account of their commitment to mathematical rigor  (the mendelians in turn trumpeted their fidelity to empiricism). By and large the biometricians rejected mendelian inheritance \citep[p85]{TheoreticalPopGenOrigins}. This may seem odd in this day and age, but remember that we are discussing a time that predates the discovery of the material basis of inheritance (chromosomes comprised of DNA), as well as the mechanism of inheritance (meiosis).

                                                                                                                    The possibility of reconciling the views of the mendelians with those of the biometricians had been considered by Yule, and Parson as early as 1902. However it is Fisher's paper, published in 1918, on ``The correlation of relatives on the supposition of Mendelian inheritance" that is widely regarded as marking the beginning of the synthesis of these two purportedly irreconcilable theories into a single theory of evolution---what we now call the \emph{modern synthesis}.

                                                                                                                   In this paper, Fisher, himself an eminent statistician, presented a mathematical model that incorporated natural selection and mendelian inheritance, and used it, amongst other things, to calculate correlations between certain traits in relatives. He argued that his model could account for published biometric data \citep{PearsonLee} which Pearson (a biometrician) had previously used to question the adequacy of Mendelian inheritance.

                                                                                                                   Fisher's use of Pearson's own data to challenge Pearson's position was seen as a coup d'\'etat of sorts. Provine \citeyearpar[p147]{TheoreticalPopGenOrigins} reports that ``Fisher's 1918 paper was well received by the few geneticists who could understand his mathematics". At any rate, there seems to have been no alarm over the remarkably strong assumptions undergirding Fisher's work. What were these assumptions?

                                                                                                                   In Fisher's model a quantitative trait  (e.g. stature, i.e. height) is under the influence of multiple genes, each with multiple allelic instantiations.  Fisher assumed that the effects of allele substitutions in an individual were constant and combined additively. This assumption entails that  the substitution of one allele for another always has the same additive effect on the value of the trait, regardless of the genetic background in which the allele substitution occurs. To be fair Fisher did discuss the possibility that the effects of gene substitutions might not combine additively. He called this condition \emph{epistacy}, a term he later revised to epistasis. Early in the paper Fisher urged his readers to treat epistasis and the effects of the environment as one might regard ``an arbitrary error introduced into the measurements", in other words, as noise. Later he returned to show how the case when epistasis is not well modeled by noise might be dealt with. Unfortunately, his treatment is rather incomplete; he limited himself to addressing ``deviations from linearity as may exist between two factors", but not more, because (amongst other things) he deemed such higher order deviations ``improbable".

                                                                                                                   Importantly, considerations of epistasis did not figure into Fisher's accounting for the biometric data of Pearson and Lee.  Fisher, in other words, accounted for this data while making \emph{extremely} strong assumptions about the effects of allele substitutions.  Nevertheless, he firmly believed that his approach was valid. In the conclusion of the paper he wrote:

                                                                                                                   \begin{quote}``Throughout this work it has been necessary not to include any avoidable complications, and for this reason the possibilities of Epistacy have only been touched upon, and small quantities of the second order have been steadily ignored. In spite of this, it is believed that the statistical properties of any features determined by a large number of Mendelian factors have been successfully elucidated". \end{quote}

                                                                                                                   Fisher's characterization of epistasis as an ``avoidable complication" to his theory betrays a confusion about the role of parsimony in scientific theorizing. He seems to have believed that a simpler \emph{explanation} is to be preferred to a more complicated one. Occam's razor, or the principle of parsimony, however applies to the \emph{assumptions undergirding an explanation}, and \emph{not} to the deductive chains of logic based upon these assumptions. To see this clearly, note that \emph{any} phenomenon can be explained by the simple statement ``God wills it". The appeal of scientific theories lies, certainly not in their comparative succinctness (scientific explanations are always longer), but in the comparative parsimony of the \emph{assumptions} involved.

                                                                                                                   As mentioned above, the extraordinary strength of Fisher's assumptions drew no protest at the time. Encouraged by the reception of his paper, Fisher continued with his particular mode of analysis \citep[p147]{TheoreticalPopGenOrigins}. In his highly influential book ``The Genetical Theory of Natural Selection", in which he set Darwin's theory of natural selection on a Mendelian foundation, Fisher \citeyearpar{fisher1930gtn} gave short shrift to epistasis. This time however his assumptions were criticized by Sewall Wright who, based on years of experimental work, was convinced of the error of assigning fitness effects to individual genes. In a review of Fisher's book Wright remarked that  the Fisherian approach \begin{quote}
                                                                                                                   ``assumes that each gene is assigned a constant value, measuring its contribution to the character of the individual (here fitness) in such a way that the sums of the contributions of all genes will equal as closely as possible the actual measures of the character in the individuals of the population. Obviously there could be exact agreement in all cases only if dominance and epistatic relationships were completely lacking. \ldots [W]ith respect to such  a character as fitness, it may safely be assumed that there are always important epistatic effects. Genes favorable in one combination, are, for example extremely likely to be unfavorable in another."\citep{wright1930rgt}
                                                                                                                   \end{quote}

                                                                                                                   Wright's protests notwithstanding, the assumption that the fitness effects of gene substitutions are independent of the genetic background in which they occur has become the orthodoxy in population genetics. This position is described on the first page of a popular introductory textbook as follows:
                                                                                                                   \begin{quote}
                                                                                                                   ``Population geneticists have achieved remarkable success by choosing to ignore the complexities of real populations and focusing on the evolution of one or a few loci at a time. \ldots The success of this approach, which has been seen in both theoretical and experimental investigations has been impressive, as I hope the reader will agree by the end of this book. The approach is not without its detractors. Years ago, Ernst Mayr mocked this approach as `bean bag genetics.' In so doing, he echoed a view held by many of the pioneers of our field that natural selection acts on highly interactive coadapted genomes whose evolution cannot be understood by considering the evolution of a few loci in isolation from all others. Although genomes are certainly coadapted, there is precious little evidence that there are strong interactions between most polymorphic alleles in natural populations. The modern view, spurred on by the rush of DNA sequence data, is that we can profitably study loci in isolation."\citep{gillespie}
                                                                                                                   \end{quote}

                                                                                                                    This ``modern view" is challenged by a small, but outspoken community of critics, who consider the practice of theoretically analyzing isolated loci to be driven by convenience and scandalously naive; see, for example, the volume by Wolf et. al \citeyearpar{episAndEvolProc}. Consider the following passage from this compilation which accounts for  Gillespie's observation that ``there is precious little evidence that there are strong interactions between most polymorphic alleles in natural populations". Paraphrasing statements made by Frankel and Shork \citeyearpar{frankel}, Templeton writes:
                                                                                                                    \begin{quote}``The subjective assessment of Frankel and Shork (1996) implies that epistasis is common, despite the numerous biases that exist against its detection. Frankel and Shork (1996) point out that the primary reason why many complex traits are not reported to have epistasis is simply that many investigators use designs and/or analytical methods that exclude epistasis. The implicit assumption in these analyses is that all the biologically important associations are to be found at the single-locus level."\citep{Templeton2000}\end{quote}

                                                                                                                    Later in the same paper, Templeton cautions that the pervasiveness of this ``implicit assumption" amounts to a community-wide neglect of epistasis because of the mathematical and statistical inconvenience it poses.
                                                                                                                    \begin{quote}``The dominance of this research paradigm has more to do with mathematical and statistical convenience than with biological reality. All that we know of biological systems---from the control of gene expression, to biochemical pathways, to developmental processes, to physiological regulation---indicates that interactions are the norm." \citep{Templeton2000}\end{quote}

                                                                                                                    Meanwhile Rice has argued that the very notion of epistasis is questionable. Not because epistasis does not exist, but because it is ubiquitous.
                                                                                                                    \begin{quote}`` `Epistasis,' like `invertebrate,' is a term that really means `everything else'. Traditionally defined as a situation in which the consequences of an allele substitution at one locus are a function of what allele is present at another locus \ldots, epistasis includes all possible ways that gene products can conspire to shape a phenotype, with the very unlikely exception of complete additivity. To name a phenomenon in this way has the curious effect of making it look like a special case, even if it is the most common situation.''\citep{rice2000}\end{quote}

                                                                                                                    \subsection{\hspace{-.2cm}Comparing Fisher's Assumptions with Holland's}

                                                                                                                   Proponents of the building block hypothesis typically regard each symbol in a genomic string as a separate gene \citep{holland75:_adapt_natur_artif_system, Goldberg:1989:GAS,Mitchell:1996:IGA}. A building block can then be (incompletely) described as a small set of closely located genes amongst which credit cannot be apportioned, i.e. a set of genes with epistatic interactions.  As building blocks are assumed to be abundant, proponents of the building block hypothesis feel justified in claiming that the building block hypothesis accommodates the existence of pervasive epistasis, and therefore rests on assumptions that are much weaker than Fisher's.

                                                                                                                   A less cozy picture however emerges if one defines a gene more in line with its definition in population genetics. The word ``gene" was coined by the Danish geneticist Wilhelm Johansen in 1909, at a time when the particulate nature of inheritance could only be surmised by observing differences between generations of phenotypes. Johansen used the word gene to stand for a \emph{unit of inheritance} passed on from parent to child in an all-or-nothing fashion. The theories of the  founders of population genetics---Fisher, Wright, Haldane---all make use of these fictitious units. Later, as the physical basis of inheritance started to become clear, molecular biologists began to identify a gene with any regulatory, transcribed, and/or other functional region of DNA.  Unfortunately this use of the term ``gene" is not consistent with its use in population genetics. In order to maintain consistency with the theoretical work of their field's founders, population geneticists typically identify a gene with  a chromosomal extent that tends to be inherited in an all-or-nothing fashion, i.e. a chromosomal extent that is short enough that it tends not to be broken up by crossover \citep[p28]{selfishGene}. This is not a strict definition, like say that of a triangle, but instead has a ``fading-out" quality that is contingent upon the expected number of crossover points and the way they tend to be distributed over the genome\footnote{Dawkins uses the word \emph{cistron} to refer to what molecular biologists call a gene \citep[p28]{selfishGene}. }.

                                                                                                                    If we focus on GAs with $n$-point crossover where $n$ is small, then in light of the above, any short schema with contiguous defining positions is a gene. Given this definition of a gene, note that any building block must be comprised  of one or more genes with above average fitness. For example, if  ****************10*1*** is a building block, then one or both of the genes  ****************1001***, and  ****************1011***  must have above average fitness. Given the above it is easy to see how the building block hypothesis and the assumptions that undergird it can be described entirely in terms of genes. Taking this perspective allows us to compare Fisher's assumptions with those of Holland.

                                                                                                                    Both Fisher and Holland assumed the existence of  a large number of genes with higher than average fitness. Fisher essentially assumed that \emph{any} collection of such genes intersects synergistically. Holland, assumed a) that synergistic intersections between small collections of such genes is common, b) that antagonistic intersections between any collection of such genes is rare, and c) that this pattern applies hierarchically. To be sure these three assumptions about the distribution of fitness \emph{are} weaker than Fisher's. However, given the extraordinary strength of Fisher's assumptions, that's not saying much.

                                                                                                                    Fisher and Holland also differ in the way they deal with the problem of sampling error.  By assuming an infinite panmictic (i.e. fully mixing)  population, Fisher dispensed with the need for the average fitness values of the alleles of each gene to be significantly different. This is because in an infinite panmictic population evolution can act on differences between these values no matter how small the differences. In other words, by making a strong assumption about the size of the evolving population Fisher was able to avoid making a strong assumption about the distribution of fitness. Because GAs used in practice tend to have small populations---typically no more than 1000 individuals---an escape of this sort is clearly not available to genetic algorithm theorists

                                                                                                                    Nevertheless, to the best of our knowledge the issue of sampling error has not been addressed by proponents of the building block hypothesis. By insisting that the building block hypothesis explains the adaptive capacity of GAs with small populations, and by making no effort to address the issue of sampling error that arises when one takes this position, these proponents are, in effect, making the assumption that each basic building block is comprised of at least one gene whose sampling fitness is likely to be so far above average that evolution will propagate this gene despite the inevitable sampling error that accompanies the evolution of small populations. The assumption that basic building blocks are abundant entails the extraordinarily strong assumption that the genes that comprise them are also abundant.

                                                                                                                     \section{The Problem with the Loose Linkage ``Problem"\label{looselinkage}}


                                                                                                                    While proponents of the building block hypothesis seem to be comfortable with the strong assumption that low-order schemata with above average sampling fitness are abundant, they are less enthusiastic about assuming abundance when it is also stipulated that the schemata must be short. In other words, these proponents are uncomfortable assuming that the defining positions of basic building blocks are ``tightly linked", i.e. close together.

                                                                                                                    Their wariness about this assumption can be traced back to Holland's contention in his seminal treatise \citep{holland75:_adapt_natur_artif_system} that the defining bits of building blocks may well be dispersed throughout the genome (i.e. loosely linked). Holland considered this to be a significant problem. To deal with it he introduced the \emph{inversion} operator which reverses the order of the bits of a randomly chosen snippet of a genome while preserving the genome's semantics \citep[p106]{holland75:_adapt_natur_artif_system}.  Holland asserted that over several generations inversion, in combination with crossover and selection, will tighten the linkage between the defining positions of schemata with above average fitness in an ``intrinsically parallel fashion"\citep[p109]{holland75:_adapt_natur_artif_system}.

                                                                                                                    The inversion operator was not found to be useful in practice \citep{davis91}, and the study of inversion did not become an active area of research. However, the loose linkage problem that inversion was supposed to solve took on a life of its own---it became a de facto explanation for poor GA performance, and has captured the attention and creative energies of a sizeable section of the GA community. A large number of algorithms with explicit ``linkage learning" schemes have been developed to deal with this perceived problem e.g. mGA and the fmGA \citep{goldberg:1989:mgamafr, Goldberg1990a, Goldberg:1993:RAO, karguptathesis}, gemGA, \citep{conf/icec/Kargupta96a,desInnov}, LLGA \citep{harik97learning,desInnov},  ECGA \citep{harik99linkage}, FDA \citep{muehlenbein:1999:fdaadf}, LFDA \citep{muehlenbein:2001:earsd}, BOA \citep{Pelikan:98aa,desInnov}, hBOA \citep{pelikan:2001:HBOABOANLS}, SEAM \citep{oai:eprints.ecs.soton.ac.uk:12006,watsonBook}.

                                                                                                                    What seems to have been overlooked in this flurry of activity is that the assumption of tight linkage between the defining positions of basic building blocks is just one of a number of strong assumptions undergirding the building block hypothesis. Consciously or otherwise,  these other assumptions---the \emph{abundance} of basic building blocks, and hierarchical synergism---have been embraced by the inventors of the algorithms listed above.

                                                                                                                     Goldberg \citeyearpar{desInnov} calls such algorithms ``competent" GAs. The implication, of course, is that the simple genetic algorithm, which lacks explicit linkage learning mechanisms, is \emph{in}competent. Driving home this point Goldberg writes: \begin{quote}``One mistaken idea that has led to controversy is the idea that simple GAs as originally designed \citep{DeJongDissertation} or their minor variants achieve the kind of robustness sought in Holland's early writing. This text puts this canard to rest; even my first text [\citep{Goldberg:1989:GAS}] went to great lengths to discuss the importance of linkage and the inadequacy of simple GAs in solving the linkage problem. Nonetheless, the field has proceeded using simple GAs as though they worked well. They do not."\citep[p55]{desInnov}\end{quote}
                                                                                                                     Sadly, Goldberg does not consider the possibility that simple GAs and minor variants thereof  continue to be used by GA practitioners because they \emph{do} ``work well", but \emph{don't} work as described in the building block hypothesis.

                                                                                                                     An unfortunate consequence of the field's preoccupation with the loose linkage ``problem" is the diversion of effort that might otherwise have been devoted to developing more viable explanations for the adaptive capacity of the simple genetic algorithm. Why, after all, would an engineer care to study the workings of an algorithm deemed the poor cousin of more ``competent" algorithms?

                                                                                                                     \section{Conclusion\label{conclusion}}

                                                                                                                     What is a genetic algorithm? Is it

                                                                                                                     \begin{description}\item[(a)] An algorithmic model of natural evolution in which selection, crossover, and mutation are iteratively applied to a population of strings, or
                                                                                                                     \item[(b)] An algorithm that implements the process described in the building block hypothesis \end{description}Proponents of the building block hypothesis regard both (a) and (b) as valid descriptions of a genetic algorithm. However, of the two descriptions, their loyalty seems to reside with description (b).  Holland, for example, writes ``the very essence of good GA design is retention of diversity, furthering exploration, while exploiting building blocks already discovered"\footnote{One has to wonder if the building block hypothesis can really be called a hypothesis if essential parts of it are incorporated into the \emph{definition} of a GA.} Most researchers in the field however readily agree with description (a), but need to be convinced that (b) is also a valid description. The  theoretical basis of the argument by which BBH proponents attempt to convince us of the validity of description (b) has been sharply criticized. Unfortunately criticism of this sort has been brushed aside by BBH proponents who continue to insist that (b) is a valid description.

                                                                                                                     ``What is it that people do when they are being innovative in a cross-fertilizing sense?", asks Goldberg.  ``Usually they are grasping at a notion---a set of good solution features---in one context, and a notion in another context and juxtaposing them, thereby speculating that the combination might be better than either notion taken individually" \citep[p5]{desInnov}. Goldberg calls the attribution to genetic algorithms of this purported process of human innovation the \emph{fundamental intuition of genetic algorithms}. One of the ways in which he attempts to justify this \emph{de facto} attribution is by reproducing a quote by the French Mathematician Jacques Hadamard, from a book entitled ``The Psychology of Invention in the Mathematical Field" : ``Indeed, it is obvious that invention or discovery be it in mathematics or anywhere else, takes place by combining ideas".

                                                                                                                     No one can deny the aesthetic appeal of the building block hypothesis---least of all computer scientists, weaned  (as we typically are) on the virtues and use of modularity. That genetic algorithms might be constructing solutions to problems much as we do---by identifying important modules, and composing them together---is a truly exciting prospect. Unfortunately, as discussed in this paper, some extraordinarily strong assumptions about the distribution of fitness have to hold true in order for this prospect to be realized.

                                                                                                                     The history of science is replete with conceptual entities that were highly popular for a while, but were later jettisoned because of the extraordinary strength of the assumptions one had to embrace. One such entity is the luminiferous (i.e. light bearing) aether, which, for much of the nineteenth century was assumed to be the medium through which light traveled. Light, like all waves, bends around objects (diffraction), changes direction upon striking a reflective surface (reflection) or entering a new medium (refraction), can be split into components with different frequencies (dispersion), and displays interference patterns.  Physicists in the nineteenth century assumed that the propagation of light,  like the propagation of all waves known at the time, requires the mechanical disturbance of some physical medium. This medium was called luminiferous aether.

                                                                                                                     On the one hand the concept of luminiferous aether proved very useful because it allowed physicists to account for phenomena like diffraction which presented serious problems for a corpuscular theory of light; on the other hand this concept presented some serious difficulties of its own. For example, the aether had to be extremely  rigid in order to support the high frequencies of light. At the same time, because it did not seem to have any observable effect on the orbits of the planets, it had to be devoid of mass and viscosity! Paradoxes like these occupied the minds of some of the finest physicists during the latter half of nineteenth century and into the early twentieth.  Ultimately, the existence of the aether was obviated by the less presumptive theories of Maxwell  (who cast light as a electromagnetic wave, \emph{not} dependent on the mechanical  properties of an aether), and Einstein (specifically his work on wave-particle duality and special relativity). The luminiferous aether, in other words, fell to Occam's razor.

                                                                                                                     Will building blocks go the way of luminiferous aether? Time will tell. At this point what we \emph{can} say  is that certain parallels between the two are unmistakeable. Both are the result of extrapolation---of the mechanical basis of wave propagation in the second case, and of the purported process underlying \emph{human} innovation in the first---and both require believers to embrace, whether consciously or otherwise, assumptions so strong that they seem almost magical.

\bibliographystyle{plainnat}

\begin{thebibliography}{50}
\providecommand{\natexlab}[1]{#1}
\providecommand{\url}[1]{\texttt{#1}}
\expandafter\ifx\csname urlstyle\endcsname\relax
  \providecommand{\doi}[1]{doi: #1}\else
  \providecommand{\doi}{doi: \begingroup \urlstyle{rm}\Url}\fi

\bibitem[Cavicchio(1970)]{cavicchio}
D.~J. Cavicchio.
\newblock \emph{Adaptive Search Using Simulated Evolution}.
\newblock PhD thesis, University Of Michigan, Ann Arbor, 1970.

\bibitem[Copeland(2004)]{Copeland}
B.~Jack Copeland.
\newblock Computation.
\newblock In Luciano Floridi, editor, \emph{The Blackwell Guide to the
  Philosopy of Computing and Information}, chapter~1, pages 3--17. Blackwell
  Publishing, 2004.

\bibitem[Davis(1991)]{davis91}
Lawrence Davis.
\newblock \emph{Handbook of genetic algorithms}.
\newblock Van Nostrand Reinhold, New York, 1991.

\bibitem[Dawkins(1999)]{selfishGene}
Richard Dawkins.
\newblock \emph{The Selfish Gene}.
\newblock Oxford University Press, 1999.

\bibitem[{De Jong}(1975)]{DeJongDissertation}
Kenneth {De Jong}.
\newblock \emph{An Analysis of the Behavior of a class of genetic adaptive
  systems}.
\newblock PhD thesis, University of Michigan, Ann Arbor, 1975.

\bibitem[Fisher(1930)]{fisher1930gtn}
RA~Fisher.
\newblock \emph{{The genetical theory of natural selection. Clarendon}}.
\newblock Oxford, 1930.

\bibitem[Forrest and Mitchell(1993)]{forrest93relative}
Stephanie Forrest and Melanie Mitchell.
\newblock Relative building-block fitness and the building-block hypothesis.
\newblock In L.~Darrell Whitley, editor, \emph{Foundations of Genetic
  Algorithms 2}, pages 109--126, San Mateo, CA, 1993. Morgan Kaufmann.

\bibitem[Frankel and Schork(1996)]{frankel}
W.~N. Frankel and N.~J. Schork.
\newblock Who's afraid of epistasis.
\newblock \emph{Nature Genetics}, 14:\penalty0 371--373, 1996.

\bibitem[Gillespie(1998)]{gillespie}
John~H. Gillespie.
\newblock \emph{Population Genetics: A Concise Guide}.
\newblock The John Hopkins University Press, 1998.

\bibitem[Goldberg et~al.(1989)Goldberg, Korb, and Deb]{goldberg:1989:mgamafr}
D.~E. Goldberg, B.~Korb, and K.~Deb.
\newblock Messy genetic algorithms: Motivation, analysis, and first results.
\newblock \emph{Complex Systems}, 3:\penalty0 493--530, 1989.

\bibitem[Goldberg(1989)]{Goldberg:1989:GAS}
David~E. Goldberg.
\newblock \emph{Genetic Algorithms in Search, Optimization \& Machine
  Learning}.
\newblock Addison-Wesley, Reading, MA, 1989.

\bibitem[Goldberg(2002)]{desInnov}
David~E. Goldberg.
\newblock \emph{The Design Of Innovation}.
\newblock Kluwer Academic Publishers, 2002.

\bibitem[Goldberg et~al.(1990)Goldberg, Deb, and Korb]{Goldberg1990a}
David~E. Goldberg, Kalyanmoy Deb, and Bradley Korb.
\newblock Messy genetic algorithms revisited: Studies in mixed size and scale.
\newblock \emph{Complex Systems}, 4\penalty0 (4):\penalty0 415--444, 1990.

\bibitem[Goldberg et~al.(1993)Goldberg, Deb, Kargupta, and
  Harik]{Goldberg:1993:RAO}
David~E. Goldberg, Kalyanmoy Deb, Hillol Kargupta, and Georges Harik.
\newblock Rapid, accurate optimization of difficult problems using fast messy
  genetic algorithms.
\newblock pages 56--64, 1993.

\bibitem[Harik(1999)]{harik99linkage}
G.~Harik.
\newblock Linkage learning via probabilistic modeling in the ecga, 1999.

\bibitem[Harik and Goldberg(1997)]{harik97learning}
Georges~R. Harik and David~E. Goldberg.
\newblock Learning linkage.
\newblock In Richard~K. Belew and Michael~D. Vose, editors, \emph{{F}oundations
  of {G}enetic {A}lgorithms 4}, pages 247--262, San Francisco, CA, 1997. Morgan
  Kaufmann.

\bibitem[Holland(1992)]{Holland92}
John~H. Holland.
\newblock Genetic algorithms.
\newblock \emph{Scientific American}, 267, 1992.

\bibitem[Holland(1975)]{holland75:_adapt_natur_artif_system}
John~H. Holland.
\newblock \emph{Adaptation in Natural and Artificial Systems: An Introductory
  Analysis with Applications to Biology, Control, and Artificial Intelligence}.
\newblock MIT Press, 1975.

\bibitem[Holland(2000)]{journals/ec/Holland00}
John~H. Holland.
\newblock Building blocks, cohort genetic algorithms, and hyperplane-defined
  functions.
\newblock \emph{Evolutionary Computation}, 8\penalty0 (4):\penalty0 373--391,
  2000.

\bibitem[Hollstien(1971)]{Hollstien}
Hollstien.
\newblock \emph{Artificial Genetic Adaptation in Computer Control Systems}.
\newblock PhD thesis, University of Michigan, Ann Arbor, 1971.

\bibitem[Kargupta(1996)]{conf/icec/Kargupta96a}
Hillol Kargupta.
\newblock The gene expression messy genetic algorithm.
\newblock pages 814--819, 1996.

\bibitem[Kargupta(1995)]{karguptathesis}
Hillol Kargupta.
\newblock \emph{SEARCH, Polynomial Complexity, and the Fast Messy Genetic
  Algorithm}.
\newblock PhD thesis, University of Illinois at Urbana-Champaign, 1995.

\bibitem[Kuhn()]{kuhn:ssr}
T.S. Kuhn.
\newblock \emph{{The Structure of Scientific Revolutions (: 1996)}}.

\bibitem[Mitchell(1996)]{Mitchell:1996:IGA}
Melanie Mitchell.
\newblock \emph{An Introduction to Genetic Algorithms}.
\newblock The MIT Press, Cambridge, MA, 1996.

\bibitem[M{\"{u}}hlenbein and Mahnig(2001)]{muehlenbein:2001:earsd}
H.~M{\"{u}}hlenbein and T.~Mahnig.
\newblock Evolutionary algorithms: From recombination to search distributions.
\newblock In Leila Kallel, Bart Naudts, and Alex Rogers, editors,
  \emph{Theoretical Aspects of Evolutionary Computing}, pages 135--173, Berlin,
  2001. Springer.

\bibitem[M{\"{u}}hlenbein and Mahnig(1999)]{muehlenbein:1999:fdaadf}
Heinz M{\"{u}}hlenbein and Thilo Mahnig.
\newblock The factorized distribution algorithm for additively decompressed
  functions.
\newblock In \emph{1999 {C}ongress on {E}volutionary {C}omputation}, pages
  752--759, Piscataway, NJ, 1999. IEEE Service Center.

\bibitem[Nix and Vose(1992)]{nix1992mga}
A.E. Nix and M.D. Vose.
\newblock {Modeling genetic algorithms with Markov chains}.
\newblock \emph{Annals of Mathematics and Artificial Intelligence}, 5\penalty0
  (1):\penalty0 79--88, 1992.

\bibitem[Okasha(2002)]{okasha2002psv}
S.~Okasha.
\newblock \emph{{Philosophy of Science: A Very Short Introduction}}.
\newblock Oxford University Press, 2002.

\bibitem[Pearson and Lee(1903)]{PearsonLee}
Karl Pearson and Alice Lee.
\newblock On the laws of inheritance in man. 1. inheritance of physical
  characters.
\newblock \emph{Biometrika}, 2:\penalty0 357--462, 1903.

\bibitem[Pelikan and Goldberg(2001)]{pelikan:2001:HBOABOANLS}
Martin Pelikan and David~E. Goldberg.
\newblock Hierarchical bayesian optimization algorithm = bayesian optimization
  algorithm + niching + local structures.
\newblock In \emph{Optimization by Building and Using Probabilistic Models
  (OBUPM) 2001}, pages 217--221, San Francisco, California, USA, 7 July 2001.

\bibitem[Pelikan et~al.(1999)Pelikan, Goldberg, and
  Cant{\'{u}}-Paz]{Pelikan:98aa}
Martin Pelikan, David~E. Goldberg, and Erick Cant{\'{u}}-Paz.
\newblock {BOA}: The {B}ayesian optimization algorithm.
\newblock In \emph{Proceedings of the Genetic and Evolutionary Computation
  Conference {GECCO}-99}, volume~I, pages 525--532. Morgan Kaufmann Publishers,
  San Fransisco, CA, 13-17 July 1999.

\bibitem[Popper(2007{\natexlab{a}})]{conjrefut}
Karl Popper.
\newblock \emph{Conjectures and Refutations}.
\newblock Routledge, 2007{\natexlab{a}}.

\bibitem[Popper(2007{\natexlab{b}})]{logicofscientificdisc}
Karl Popper.
\newblock \emph{The Logic Of Scientific Discovery}.
\newblock Routledge, 2007{\natexlab{b}}.

\bibitem[Provine(2001)]{TheoreticalPopGenOrigins}
William~B. Provine.
\newblock \emph{The Origins of Theoretical Population Genetics}.
\newblock The University of Chicago Press, Chicago, 2001.

\bibitem[Pr{\"u}gel-Bennett and Shapiro(1994)]{prugelbennett1994aga}
A.~Pr{\"u}gel-Bennett and J.L. Shapiro.
\newblock {Analysis of genetic algorithms using statistical mechanics}.
\newblock \emph{Physical Review Letters}, 72\penalty0 (9):\penalty0 1305--1309,
  1994.

\bibitem[Rattray(1996)]{RattrayThesis}
Magus~L. Rattray.
\newblock \emph{Modelling the Dynamics of a Genetic Algorithm Using Statistic
  Mechanics}.
\newblock PhD thesis, Computer Science Department, University of Manchester,
  1996.

\bibitem[Reeves and Rowe(2003)]{reeves2003gap}
C.R. Reeves and J.E. Rowe.
\newblock \emph{{Genetic Algorithms: Principles and Perspectives: a Guide to GA
  Theory}}.
\newblock Kluwer Academic Publishers, 2003.

\bibitem[Rice(2000)]{rice2000}
Sean~H. Rice.
\newblock The evolution of developmental interactions.
\newblock In  \citet{episAndEvolProc}.

\bibitem[Shapiro(2001)]{shapiro:2001:smtga}
J.~L. Shapiro.
\newblock Statistical mechanics theory of genetic algorithms.
\newblock In Leila Kallel, Bart Naudts, and Alex Rogers, editors,
  \emph{Theoretical Aspects of Evolutionary Computing}, pages 87--108, Berlin,
  2001. Springer.

\bibitem[Simon(1969)]{Simon69}
H.~Simon.
\newblock \emph{The Sciences of the Artificial}.
\newblock The M.I.T. Press, 1969.

\bibitem[Templeton(2000)]{Templeton2000}
Alan~R. Templeton.
\newblock Epistasis and complex traits.
\newblock In Jason~B. Wolf, Edmund~D. Brodie, and Michael~J. Wade, editors,
  \emph{Epistasis and the Evolutionary Process}. Oxford University Press, 2000.

\bibitem[Tversky and Kahneman(1974)]{tversky1974juu}
A.~Tversky and D.~Kahneman.
\newblock {Judgment under Uncertainty: Heuristics and Biases}.
\newblock \emph{Science}, 185\penalty0 (4157):\penalty0 1124--1131, 1974.

\bibitem[Vose(1993)]{Vose1993}
Michael~D. Vose.
\newblock Modeling simple genetic algorithms.
\newblock In L.~Darrell Whitley, editor, \emph{Foundations of Genetic
  Algorithms 2}, pages 63--73, San Mateo, 1993. Morgan Kaufmann.

\bibitem[Vose(1999)]{vose:1999:sgaft}
Michael~D. Vose.
\newblock \emph{The simple genetic algorithm: foundations and theory}.
\newblock MIT Press, 1999.

\bibitem[Vose and Liepins(1991)]{VoseLiepins91}
Michael~D. Vose and Gunar~E. Liepins.
\newblock Punctuated equlibria in genetic search.
\newblock \emph{Complex Systems}, 5\penalty0 (1):\penalty0 31--44, February
  1991.

\bibitem[Watson(2002)]{oai:eprints.ecs.soton.ac.uk:12006}
Richard~A. Watson.
\newblock \emph{Compositional Evolution: Interdisciplinary Investigations in
  Evolvability, Modularity, and Symbiosis}.
\newblock PhD thesis, April~01 2002.

\bibitem[Watson(2006)]{watsonBook}
Richard~A. Watson.
\newblock \emph{Compositional Evolution: The Impact of Sex, Symbiosis and
  Modularity on the Gradualist Framework of Evolution}.
\newblock The MIT Press, 2006.

\bibitem[Wolf et~al.(2000)Wolf, III, and Wade]{episAndEvolProc}
Jason~B. Wolf, Edmund D.~Brodie III, and Michael~J. Wade, editors.
\newblock \emph{Epistasis And the Evolutionary Process}.
\newblock Oxford University Press, 2000.

\bibitem[Wright et~al.(2003)Wright, Vose, and Rowe]{conf/gecco/WrightVR03}
Alden~H. Wright, Michael~D. Vose, and Jonathan~E. Rowe.
\newblock Implicit parallelism.
\newblock In \emph{GECCO}, 2003.

\bibitem[Wright(1930)]{wright1930rgt}
S.~Wright.
\newblock {Review of The Genetical Theory of Natural Selection, by RA Fisher}.
\newblock \emph{Journal of Heredity}, 21:\penalty0 80--87, 1930.

\end{thebibliography}

\end{document}